# STOP: Spatiotemporal Orthogonal Propagation for Weight-Threshold-Leakage Synergistic Training of Deep Spiking Neural Networks

Haoran Gao, *Student Member, IEEE*, Xichuan Zhou, *Senior Member, IEEE*, Yingcheng Lin, Min Tian, *Member, IEEE*, Liyuan Liu, *Member, IEEE* and Cong Shi*, *Member, IEEE*

***Abstract*—The prevailing of artificial intelligence-of-things calls for higher energy-efficient edge computing paradigms, such as neuromorphic agents leveraging brain-inspired spiking neural network (SNN) models based on spatiotemporally sparse binary spikes. However, the lack of efficient and high-accuracy deep SNN learning algorithms prevents them from practical edge deployments at a strictly bounded cost. In this paper, we propose the spatiotemporal orthogonal propagation (STOP) algorithm to tackle this challenge. Our algorithm enables fully synergistic learning of synaptic weights as well as firing thresholds and leakage factors in spiking neurons to improve SNN accuracy, in a unified temporally-forward trace-based framework to mitigate the huge memory requirement for storing neural states across all time-steps in the forward pass. Characteristically, the spatially-backward neuronal errors and temporally-forward traces propagate orthogonally to and independently of each other, substantially reducing computational complexity. Our STOP algorithm obtained high recognition accuracies of 94.84%, 74.92%, 98.26% and 77.10% on the CIFAR-10, CIFAR-100, DVS-Gesture and DVS-CIFAR10 datasets with adequate deep convolutional SNNs of VGG-11 or ResNet-18 structures. Compared with other deep SNN training algorithms, our method is more plausible for edge intelligent scenarios where resources are limited but high-accuracy in-situ learning is desired.**

***Index Terms*—neuromorphic computing, spiking neural network (SNN), backpropagation through time (BPTT), spatiotemporal backpropagation (STBP), synergistic learning**

## I. Introduction

Deep learning [1]-[3] have found a tremendous amount of applications across a widespread range of intelligent scenarios including assisted driving, security monitor, healthcare wearables, industrial robots, disaster prediction and so forth. However, conventional deep learning models relying on the continuously-valued artificial neural network (ANN) involve computationally intensive tensor multiplications. They are thus not suitable for ubiquitous edge devices in the era of artificial intelligence-of-things, where massive sensory data call for in-situ and adaptive processing under stringent energy, cost and latency budgets [4]. In contrast, the human brain is capable of conducting complex cognitive tasks while dissipating a bulb-level power as low as 20 W [5]. Such amazing energy efficiency arises in the mechanism of the cortical substrate that utilizes spatiotemporally sparse electric pulses (i.e., spikes) to exchange information among neurons. Inspired by this observation, the brain-mimicry spiking neural network (SNN) models [6]-[8] have emerged with substantially higher energy efficiency and lower computation complexity than their ANN counterparts if dedicate neuromorphic silicon is available [9]-[19].

Early on, SNNs are primarily trained by the spike-timing dependent plasticity (STDP) rule [20], [21], which modifies synaptic weights of neurons according to their pre- and post-synaptic spike activities, optionally modulated by a global reward signal [22]. Despite its high biological plausibility and computational simplicity, the STDP rule usually results in limited recognition accuracies for SNNs compared with their ANN rivals whose weights are precisely adjusted via error gradient descent [23]. To bridge the performance gap between SNNs and ANNs, the ANN-SNN conversion method appears [24]-[30]. It treats the spike rate of a spiking neuron as an approximation of the continuous-valued activation of an artificial neuron. To obtain a high-accuracy SNN, a structurally equivalent ANN is firstly trained by the standard error backpropagation (BP) process, and then the learned weights are transferred to the SNN. However, such conversion method cannot leverage the precise timing of individual spikes and often requires dozens to hundreds of time-steps to reach an acceptable level of activation approximation, which severely compromises SNN computation efficiency and incurs huge latencies on edge devices. Moreover, it is difficult and expensive to implement such conversion in an online manner for edge neuromorphic hardware processors. In recent years, BP-based techniques have been investigated for direct training of deep SNNs [31]-[32]. To overcome the non-continuity of spike activities, smooth surrogate gradients have been adopted to approximate the derivative of the non-differentiable spike firing function [33]. To handle the rich temporal dynamics in SNNs, the standard BP has to be extended into the temporal dimension, and the errors need to propagate through both space

This work was funded in part by the National Natural Science Foundation of China under Grant No. 62334008, in part by the Opening Fund of Artificial Intelligence Key Laboratory of Sichuan Province under Grant No. 2023RYY07, and in part by the by the Fundamental Research Funds for the Central Universities under Grant No. 2024CDJXY020.

Haoran Gao, Xichuan Zhou, Yingcheng Lin and Cong Shi are with the School of Microelectronics and Communication Engineering, Chongqing University, Chongqing 400044, China. (Corresponding author: Cong Shi, e-mail: shicong@cqu.edu.cn)

Min Tian is with Chongqing United Microelectronics Center Co. Ltd, Chongqing 401332, China.

Liyuan Liu is with the State Key Laboratory for Superlattices and Microstructures, Institute of Semiconductors, Chinese Academy of Sciences, Beijing 100083, China.

(i.e., the layer depth) and time (i.e., the time-steps). The well-known backpropagation through time (BPTT) algorithm [34], originally proposed to train recurrent ANNs, can handle both explicit (due to recurrent synaptic weights) and implicit (due to inherent dynamics of spiking neurons) temporal dependencies in SNNs [35]. Especially, for common image classification tasks that do not necessarily require SNNs to contain recurrent synaptic connections, the BPTT algorithm simplifies to the spatio-temporal backpropagation (STBP) rule for non-recurrent SNNs [36]. Compared to conversion-based methods, the BPTT and STBP algorithms can achieve comparable SNN accuracies with much fewer time-steps, greatly facilitating a low latency.

Nevertheless, the BPTT/STBP algorithms [34], [36] and their variants [37]-[40] bring forth a high memory cost for error BP. One has to store neural states (e.g., the internal membrane potential, the firing status, etc.) of all neurons across all time-steps in the feedforward pass, for inquiries during the later backward pass [41]. Moreover, other model parameters like firing thresholds and neural leakages in individual spiking neurons need to be manually tuned, which imposes significant challenges on optimizing deep SNN accuracies.

Several techniques have emerged to alleviate the above disadvantages of the BTTT/STBP algorithms. One solution to lower the memory complexity is leveraging eligibility traces to delegate the temporal part of the error gradients with respect to synaptic weights [42]-[44]. Since the traces can be computed online in a forward manner along time-steps, the huge storage requirement for neuronal state backup no longer exists. During the backward pass, the output errors only need to backpropagate in the spatial domain. A more aggressive approach to mitigate memory-intensive temporal backpropagation is to completely ignore temporal dependencies between different time points [43], [45]-[47]. But they involve other complicated mechanisms or poses specialized constraints on neural dynamics to maintain competent accuracies. To further improve SNN accuracy, a synergistic learning strategy is employed to concurrently train both synaptic weights and other key parameters (i.e., the firing thresholds [48]-[50] and the neural leakages [50]-[51]) under the STBP framework, acting as an alternative to the tedious manual parameter tuning process for accuracy improvement. However, these methods do not employ trace-based temporally forward propagation paradigm, thereby once again suffering the high memory complexity as in the original STBP algorithm.

This paper aims to achieve both high accuracy and high memory efficiency for non-recurrent deep SNN training. To this end, we propose the spatiotemporal orthogonal propagation (STOP) algorithm, which leverages a unified temporally-forward trace-based framework to implement fully weight-threshold-leakage (WTL) synergistic learning. The synaptic weights, the firing thresholds and the neural leakages all have their own traces, which are computed and forward propagated in the temporal domain, orthogonal to the spatial error backpropagation flow. In other words, the temporal traces and the spatial gradients are propagated independently of each other. On top of that, our high-accuracy algorithm does not necessarily exploit sophisticated spiking neuron models [39], [42], [45], auxiliary decision/loss functions [44], [45], [46], [48], [49] or mandatory normalizations [36], [37], [38], [46], making it very attractive to edge systems with quite limited memory and computing resources. The main contributions of this article are two folds:

- We propose the STOP algorithm to enable high-accuracy and memory-efficient fully synergistic learning of deep SNNs within a unified temporally-forward trace-based framework. To the best of our knowledge, this is the first time that firing thresholds and neural leakages other than synaptic weights are learned in a temporally forwarding manner.
- We have conducted extensive experiments to validate the efficacy of the proposed STOP algorithm on static image datasets CIFAR-10, CIFAR-100, as well as more challenging neuromorphic datasets DVS-Gesture and DVS-CIFAR10, with intermediately deep convolutional SNN structures.

The rest of this paper is organized as follows. Section II describes the foundations including the spiking neuron model, input encoding, output decoding and the loss function adopted in our work. Section III presents the proposed STOP algorithm for memory-efficient deep feedforward SNN training. The high accuracy of our algorithm is validated in Section IV, along with related work comparisons and discussions. Section V concludes this paper and indicates future directions.

## II. SNN Preliminaries

### A. Leaky Integrate-and-Fire Spiking Neuron

The most commonly used spiking neuron model in SNNs is the leaky integrate-and-fire (LIF) neuron, characteristic of a good balance between biological plausibility and computational efficiency [6]. Fig. 1(a) shows a feedforward SNN composed of LIF neurons, whose internal structure and temporal dynamics are illustrated in Fig. 1(b). The LIF spiking neuron continuously integrates weighted incoming (i.e., presynaptic) spikes onto its membrane potential via a set of synapses. Once its membrane potential crosses the firing threshold, the neuron issues an outgoing (i.e., postsynaptic) spike, and immediately resets itself by subtracting the threshold from the membrane potential. Meanwhile, the membrane potential is exponentially leaking all the time. More particularly, the temporal dynamics of an LIF neuron in the discrete time domain can be formulated as:

$$U_j^l[t] = \alpha^l (U_j^l[t-1] - \theta_j^l s_j^l[t-1]) + \sum_i w_{ji}^l s_i^{l-1}[t] \tag{1}$$

$$s_j^l[t] = H(U_j^l[t] - \theta_j^l) \tag{2}$$

where $t$ represents the discrete time-step, $U_j^l$, $s_j^l$, and $\theta_j^l$ ($\theta_j^l > 0$) are the membrane potential, the fired binary spike, and the firing threshold of neuron $j$ in layer $l$, respectively, while $\alpha^l$ ($0 \le \alpha^l \le 1$) means the leakage factor shared by all neurons in layer $l$, $w_{ji}^l$ denotes the weight of the synapse connecting this neuron and a particular presynaptic neuron indexed as $i$ in the preceding layer, and $H(\cdot)$ delegates the firing action based on the Heavyside step function: $H(x) = 1$ or 0 when $x \ge 0$ or $x < 0$, respectively. The term $-\theta_j^l s_j^l[t-1]$ in Eq. (1) corresponds to reset-by-subtraction.

To overcome the non-differentiability of the firing function $H(x)$ when back propagating errors from $s_j^l[t]$ down to $U_j^l[t]$

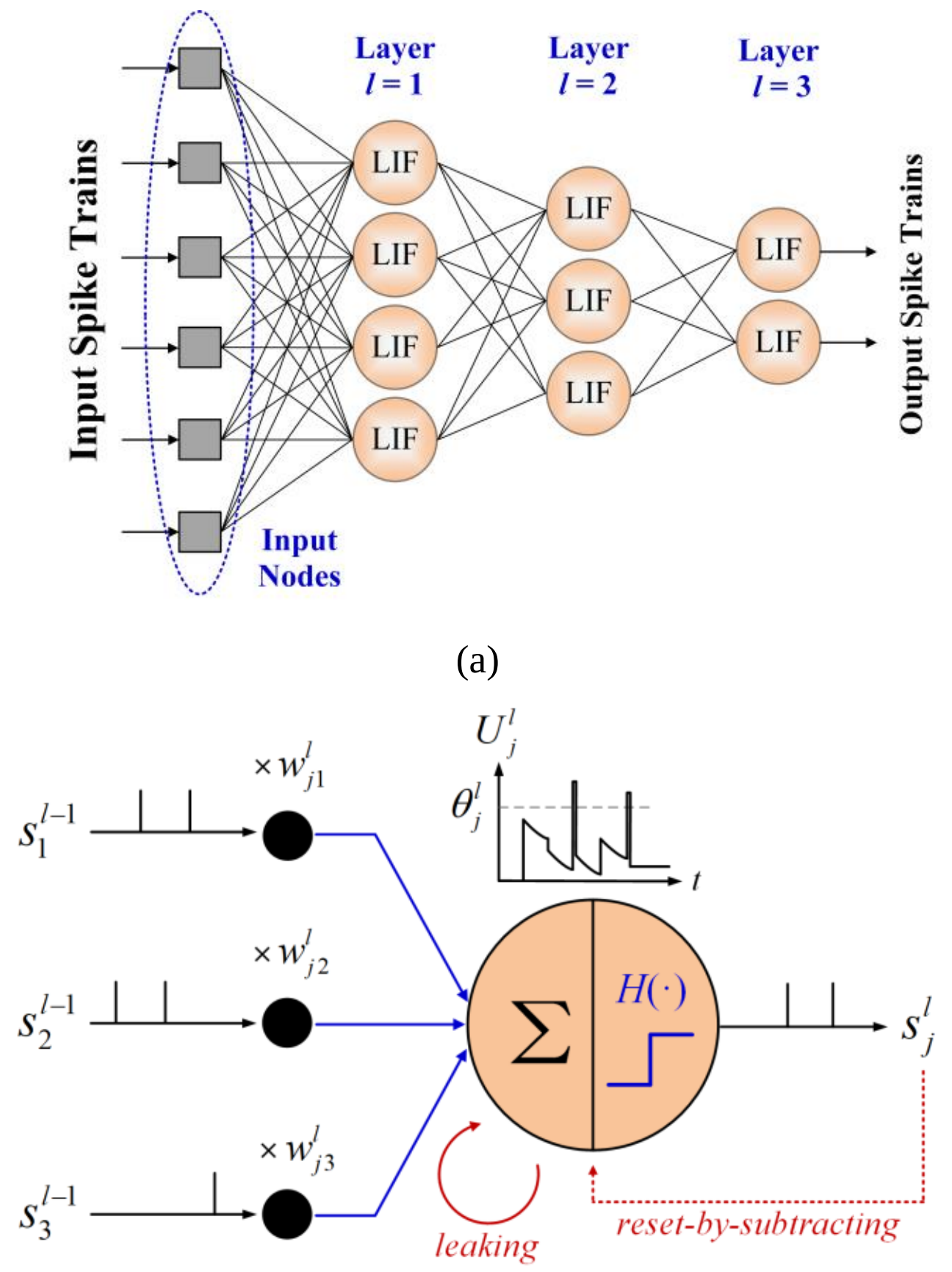

Fig. 1. (a) A feedforward SNN consisting of LIF neurons. (b) Temporal dynamics of an LIF neuron.

during training, a smooth *surrogate gradient* function $\varphi_{\mathrm{SG}}(\cdot)$ has to be employed to approximate the derivative as:

$$\frac{\partial s_j^l[t]}{\partial U_j^l[t]} \approx \varphi_{\mathrm{SG}}(U_j^l[t]-\theta_j^l) \tag{3}$$

Some common forms of $\varphi_{\mathrm{SG}}(\cdot)$ are given in [36], and Fig. 2 plots two typical exemplars.

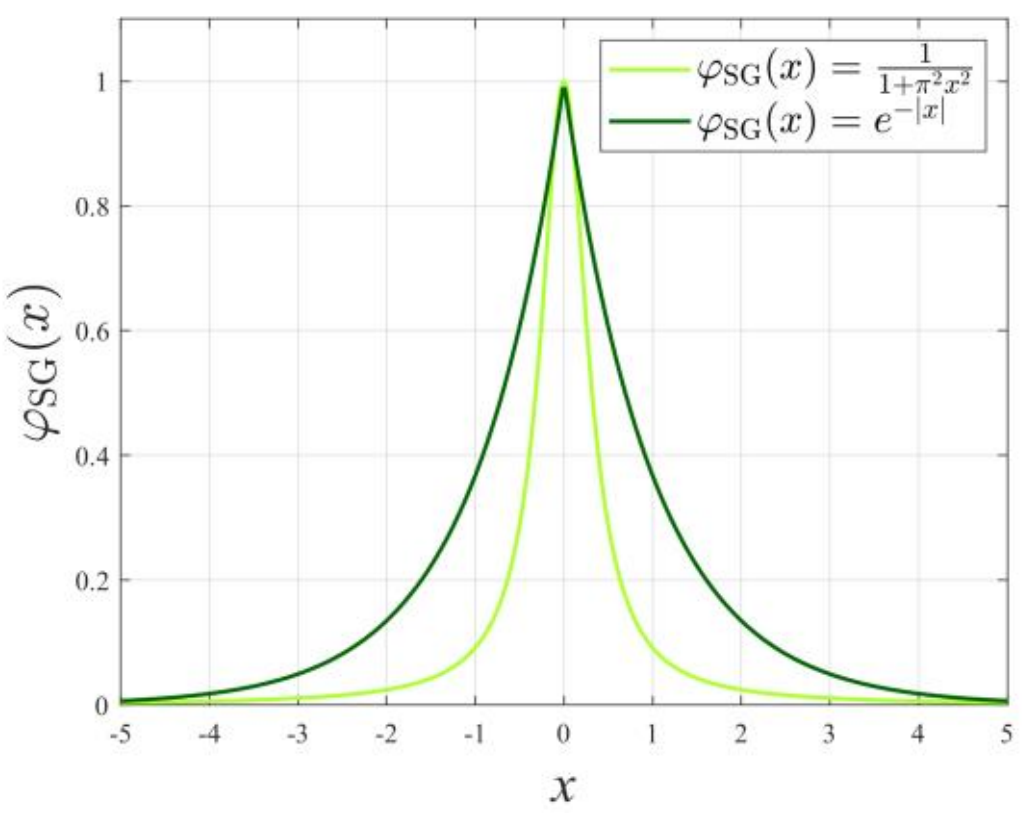

Fig. 2. Two typical surrogate gradient functions [36].

### B. Input Encoding

There are three mainstream methods for encoding continuous values (e.g., the pixel data in an image sample) before they are fed to SNNs. The rate coding method [52] converts a data value to a Poisson-distributed spike train at a rate proportional to the value, gaining robustness at higher computation overheads. The temporal coding method, on the contrary, transforms the value into a single spike occurring at a time inversely proportional to the value, trading off robustness for computational efficiency [52]. Recently, direct coding has dominated the regime of high-accuracy deep SNN training [38]. It rescales input values to the continues range of [0,1] and treats them as fractional spikes (in contrast to the ordinary binary spikes) repeatedly entering SNN at every time-step. In this work, we adopt the direct encoding.

### C. Output Decoding

In an object classification task, each neuron in the output layer is pre-allocated uniquely to a particular object category. During inference, the neuron firing the most spikes indicates the classification result of the input sample.

### D. Loss Function

The loss function reflects the deviations of actual SNN outputs from desired ones. The output loss is used to compute the errors of neurons for synaptic weight updates. The total loss $E^*$ is the aggregation of instantaneous losses $E[t]$ across all time-steps:

$$E^* = \sum_t E[t] \tag{4}$$

where $E[t]$ usually takes the form of either the cross-entropy (CE) loss or the mean squared error (MSE) loss:

CE: $$E[t] = -\sum_j s_j^d \log(\mathrm{softmax}(\mathbf{s}^L[t])_j) \tag{5}$$

MSE: $$E[t] = \frac{1}{2}\sum_j \| s_j^L[t] - s_j^d \|^2 \tag{6}$$

where $$\mathrm{softmax}(\mathbf{s}^L[t])_j = \frac{\exp(s_j^L[t])}{\sum_k \exp(s_k^L[t])} \tag{7}$$

and $\mathbf{s}^L[t] = (s_1^L[t], s_2^L[t], \ldots)^T$ denotes the actually fired spikes from neurons in the output layer $l = L$ at time $t$, $\mathbf{s}^d = (s_1^d, s_2^d, \ldots)^T$ the corresponding desired one-hot output spikes (i.e., for each $j$, $s_j^d = 1$ if the category label of the training sample equals to $j$, and $s_j^d = 0$ otherwise).

## III. The Proposed STOP Learning Algorithm

### A. Basic Concept of STOP Learning

The primary motivation and basic principle of the proposed STOP learning method is to factorize the gradient of the SNN output loss into two types of gradient components that can be computed independently in the spatial domain and in the temporal domain, respectively. Each such component is allow to propagate (i.e., calculated iteratively) only in its own domain without intruding the other one. As a result, the computational efficiency of the learning process can be largely improved. Specifically, to achieve this goal, we start from the total output loss $E^*$, and investigate its gradient with respect to a generic parameter, say $v_j^l$ in neuron $j$ of layer $l$:

$$\Delta v_j^l = \frac{\partial E^*}{\partial v_j^l} = \sum_t \frac{\partial E[t]}{\partial v_j^l} = \sum_t \frac{\partial E[t]}{\partial U_j^l[t]} \frac{\partial U_j^l[t]}{\partial v_j^l} \tag{8}$$

Here, we define

$$\delta_j^l[t] = \frac{\partial E[t]}{\partial U_j^l[t]} \tag{9}$$

as the *instantaneous neuron error* for a single time-step $t$. It reflects a direct impact of $U_j^l[t]$ on $E[t]$ of the same time, and can be easily obtained by applying the standard BP chain rule. At each time-step $t$, it is back propagated from $E[t]$ to $U_j^l[t]$ purely in the spatial domain, just as done in conventional non-spiking neural networks, except for the introduction of a surrogate function to evaluate Eq. (3). Therefore, the neuron error $\delta_j^l[t]$ can be regarded as the spatial gradient component.

On the other side, we define:

$$\tilde{v}_j^l[t] = \frac{\partial E[t]}{\partial v_j^l} \tag{10}$$

which is more computationally complex than $\delta_j^l[t]$. Because $v_j^l[t]$ impacts $U_j^l[t]$ not only directly at the current time-step $t$, but also indirectly through those past states $U_j^l[t-1]$, $U_j^l[t-2]$, …, $U_j^l[1]$, as $U_j^l[t]$ evolves along time in Eq. (1) and thus depends on its historic values. Therefore, the computation of $\tilde{v}_j^l[t]$ in Eq. (10) inevitably involves the temporal domain. However, if we can evaluate it iteratively and forward along the temporal axis without stretching into the spatial domain, as to satisfy the basic STOP principle above, the calculation of this component can be much less sophisticated. In that case, $\tilde{v}_j^l[t]$ is called a temporal trace and acts as the temporal gradient component. Actually, the temporally-forward iterative form of such trace can be easily uncovered by simultaneously differentiating both sides of Eq. (1) with respect to the parameter concerned.

In the remainder of this section, we solve the spatial neuron errors $\delta_j^l[t]$, and figure out the temporal trace forms of synaptic weights, firing thresholds and leakage factors to implement their learning updates. Finally, we combine them all to complete our temporally-forward trace-based STOP synergistic learning framework, and conduct a detailed complexity analysis on it.

### B. Neuron Error Computation

We solve the instantaneous neuron errors $\delta_j^l[t]$ of all neurons via a spatial BP procedure. For the output layer $l = L$, by taking into account Eq. (3) and (9), we obtain:

$$\delta_j^L[t] = \frac{\partial E[t]}{\partial U_j^L[t]} = \frac{\partial E[t]}{\partial s_j^L[t]} \frac{\partial s_j^L[t]}{\partial U_j^L[t]} = \frac{\partial E[t]}{\partial s_j^L[t]} \varphi_{\text{SG}}(U_j^L[t] - \theta_j^L) \tag{11}$$

Specifically, for the usage of CE loss in Eq. (5),

$$\frac{\partial E[t]}{\partial s_j^L[t]} = (\text{softmax}(\mathbf{s}^L[t]))_j - s_j^d) \tag{12}$$

while for the MSE loss in Eq. (6),

$$\frac{\partial E[t]}{\partial s_j^L[t]} = s_j^L[t] - s_j^d \tag{13}$$

Once the errors $\delta_k^{l+1}[t]$ of all neurons in layer $l + 1$ have been obtained, they can be spatially backpropagated to the lower layer $l$ to compute the errors $\delta_j^l[t]$:

$$\delta_j^l[t] = \frac{\partial E[t]}{\partial U_j^l[t]} = \sum_k \frac{\partial E[t]}{\partial U_k^{l+1}[t]} \frac{\partial U_k^{l+1}[t]}{\partial s_j^l[t]} \frac{\partial s_j^l[t]}{\partial U_j^l[t]} = \sum_k \delta_k^{l+1}[t] \frac{\partial U_k^{l+1}[t]}{\partial s_j^l[t]} \frac{\partial s_j^l[t]}{\partial U_j^l[t]} = \sum_k \delta_k^{l+1}[t] \frac{\partial U_k^{l+1}[t]}{\partial s_j^l[t]} \varphi_{\text{SG}}(U_j^l[t] - \theta_j^l) \tag{14}$$

And based on Eq. (1), we have:

$$\frac{\partial U_k^{l+1}[t]}{\partial s_j^l[t]} = w_{kj}^{l+1} \tag{15}$$

By substituting Eq. (15) into (14), we finally acquire:

$$\delta_j^l[t] = \sum_k \delta_k^{l+1}[t] w_{kj}^{l+1} \varphi_{\text{SG}}(U_j^l[t] - \theta_j^l) \tag{16}$$

### C. Trace-based Weight Learning

To update a particular synaptic weight $w_{ji}^l$, we replace the generic symbol $v_j^l[t]$ in Eq. (8) with the dedicate variable $w_{ji}^l$, and follow the definitions and conventions in Eq. (9), (10) to obtain:

$$\Delta w_{ji}^l = \frac{\partial E^*}{\partial w_{ji}^l} = \sum_t \frac{\partial E[t]}{\partial U_j^l[t]} \frac{\partial U_j^l[t]}{\partial w_{ji}^l} = \sum_t \delta_j^l[t] \tilde{w}_{ji}^l[t] \tag{17}$$

where the neuron error can be computed backward in the spatial domain through Eq. (11) and (16), and the iterative dependency of the temporal trace $\tilde{w}_{ji}^l[t]$ can be revealed by differentiating both sides of Eq. (1) with respect to $w_{ji}^l$:

$$\begin{aligned}\frac{\partial U_j^l[t]}{\partial w_{ji}^l} &= \alpha^l \left(\frac{\partial U_j^l[t-1]}{\partial w_{ji}^l} - \theta_j^l \frac{\partial s_j^l[t-1]}{\partial U_j^l[t-1]} \frac{\partial U_j^l[t-1]}{\partial w_{ji}^l}\right) + s_i^{l-1}[t] \\ &= \alpha^l \frac{\partial U_j^l[t-1]}{\partial w_{ji}^l} - \alpha^l \theta_j^l \varphi_{\text{SG}}(U_j^l[t-1] - \theta_j^l) \frac{\partial U_j^l[t-1]}{\partial w_{ji}^l} + s_i^{l-1}[t]\end{aligned} \tag{18}$$

Recall the trace definition $\tilde{w}_{ji}^l[t]$, and it transforms Eq. (18) into:

$$\tilde{w}_{ji}^l[t] = \alpha^l \tilde{w}_{ji}^l[t-1] \underbrace{-\alpha^l \theta_j^l \varphi_{\text{SG}}(U_j^l[t-1] - \theta_j^l) \tilde{w}_{ji}^l[t-1]}_{\text{illusory component}} + s_i^{l-1}[t] \tag{19}$$

which implies that $\tilde{w}_{ji}^l[t]$ is forward propagated in the temporal domain. However, it is noteworthy that the existence of the illusory component labeled in Eq. (19) relies on the surrogate gradient $\varphi_{\text{SG}}(\cdot)$ that suggests an *illusory fractional spike* (i.e., with a value in between 0 and 1) to have always been fired at the previous time-step and caused partially reset, as the smooth surrogate curves in Fig. 2 do not quickly drop to zero when $x < 0$. Such intrusions of the spatial components into the temporal domain violates our basic principle that the spatial and temporal gradients must only propagate in their own domains without interweaving with each other, and hence substantially increase computational overheads. Moreover, temporal propagation of such illusory components performs almost ineffectively and even sometimes detrimentally to SNN accuracy improvement, as will be experimentally validated later in Section IV-B. As a consequence, we preclude the illusory spatial component from the temporal propagation, and eventually reduce Eq. (19) to be:

$$\tilde{w}_{ji}^l[t] = \alpha^l \tilde{w}_{ji}^l[t-1] + s_i^{l-1}[t] \tag{20}$$

Indeed, the temporally-forward spike-driven variable $\tilde{w}_{ji}^l[t]$ just plays the role of an eligibility trace [42]. It is confirmed that in the biological brain cortex, various traces serve as certain types of memorizations allowing for retention of temporally past events, and the traces can tell the impacts of those events on future neural system states [42]. Obviously, this trace initializes to $\tilde{w}_{ji}^l[0] = 0$, since $U_j^l[t] = 0$ whenever $t \leq 0$. Moreover, notice that the update of $\tilde{w}_{ji}^l[t]$ in Eq. (20) is only driven by spikes from corresponding neuron $i$ in the preceding layer $l$ - 1 and it is regardless of the neuron index $j$ in layer $l$. These suggests that we only need to track one trace for each neuron $i$ of layer $l$ - 1, and share it with all neurons $j$ in layer $l$.

*Short summary for weight learning:* Eq. (11), (16), (17), (20) constitute the trace-based synaptic weight learning rule. For an input training sample, at each time-step $t$ = 1, 2, ...$T$, the neuron errors (spatial gradients) $\delta_j^l[t]$ are spatially back-propagated from output layer $l = L$ down to the first layer $l = 1$ according to Eq. (11), (16), while the weight-related traces (temporal gradients) $\tilde{w}_{ji}^l[t]$ are temporally forward-propagated from time-step $t$ – 1 to $t$ via Eq. (20). After the training sample presentation is over, each weight is updated following Eq. (17). Section III-F will describe this procedure more formally with pseudocodes.

## D. Trace-based Threshold Learning

Like synaptic weights, the firing thresholds of neurons can also be learned based on its corresponding temporally forward-propagated traces. One may obtain the following formulation by imitating Eq. (17) as:

$$\begin{aligned}\Delta\theta_j^l &= \frac{\partial E^*}{\partial \theta_j^l} = \sum_t \frac{\partial E[t]}{\partial U_j^l[t]} \frac{\partial U_j^l[t]}{\partial \theta_j^l} \\ &= \sum_t \delta_j^l[t] \tilde{\theta}_j^l[t]\end{aligned} \tag{21}$$

Nevertheless, slightly differing from the synaptic weight $w_{ji}^l$, which affects the output loss $E[t]$ only through the unique path serially across the node $U_j^l[t]$ in Eq. (1) and the node $s_j^l[t]$ in Eq. (2), the threshold $\theta_j^l$ additionally contributes to $E[t]$ along another path via $s_j^l[t]$ yet bypassing $U_j^l[t]$, as paved by the term $-\theta_j^l$ in Eq. (2). Therefore, Eq. (21) has to be modified as below to incorporate the influence of that extra path:

$$\Delta\theta_j^l = \sum_t \left(\delta_j^l[t] \tilde{\theta}_j^l[t] + \frac{\partial E[t]}{\partial s_j^l[t]} \frac{\partial s_j^l[t]}{\partial \theta_j^l}\right) \tag{22}$$

In view of the symmetric roles of $U_j^l[t]$ and $-\theta_j^l$ for the $s_j^l[t]$ generation in Eq. (2), we can get:

$$\begin{aligned}\frac{\partial E[t]}{\partial s_j^l[t]} \frac{\partial s_j^l[t]}{\partial \theta_j^l} &= -\frac{\partial E[t]}{\partial s_j^l[t]} \frac{\partial s_j^l[t]}{\partial(-\theta_j^l)} = -\frac{\partial E[t]}{\partial s_j^l[t]} \frac{\partial s_j^l[t]}{\partial U_j^l[t]} \\ &= -\frac{\partial E[t]}{\partial U_j^l[t]} = -\delta_j^l[t]\end{aligned} \tag{23}$$

Substituting Eq. (23) into (22) leads to:

$$\begin{aligned}\Delta\theta_j^l &= \sum_t \left(\delta_j^l[t] \tilde{\theta}_j^l[t] - \delta_j^l[t]\right) \\ &= \sum_t \delta_j^l[t] (\tilde{\theta}_j^l[t] - 1)\end{aligned} \tag{24}$$

To solve the temporal trace $\tilde{\theta}_j^l[t]$, we differentiate both sides of Eq. (1) with respect to $\theta_j^l$, and once again drop the illusory component evoked by the surrogate gradient. We finally have:

$$\frac{\partial U_j^l[t]}{\partial \theta_j^l} = \alpha^l \left(\frac{\partial U_j^l[t-1]}{\partial \theta_j^l} - s_j^l[t-1]\right) \tag{25}$$

Substituting the $\tilde{\theta}_j^l[t]$ definition into Eq. (25) results in:

$$\tilde{\theta}_j^l[t] = \alpha^l (\tilde{\theta}_j^l[t-1] - s_j^l[t-1]) \tag{26}$$

This indicates that the threshold-related trace $\tilde{\theta}_j^l[t]$ can also be computed online in a temporally-forward manner as the weight-related trace $\tilde{w}_{ji}^l[t]$ does in Eq. (20), expect the update of $\tilde{\theta}_j^l[t]$ is driven by postsynaptic spikes of the neuron and decays with the leakage factor of the neuronal membrane potential. The initial values of $\tilde{\theta}_j^l[0]$ are 0, too.

*Short summary for threshold learning:* Eq. (11), (16), (21), (26) constitute the trace-based firing threshold learning rule. The steps for threshold updates resemble the summary for weight learning in the previous subsection. Section III-F will leverage pseudocodes to describe these steps more precisely.

Besides, the firing thresholds have to be truncated above 0 to satisfy its non-negative nature each time they execute updates.

### *E. Trace-based Leakage Learning*

Although the leakage factor is shared by all neurons in one layer, its update amount has to be evaluated over individual neurons and then averaged. To obtain each neuron's leakage factor change amount $\Delta\alpha_j^l$ via a temporally-forwarded trace, we likewise form a learning equation like Eq. (17):

$$\Delta\alpha_j^l = \frac{\partial E^*}{\partial \alpha^l} = \sum_t \frac{\partial E[t]}{\partial U_j^l[t]} \frac{\partial U_j^l[t]}{\partial \alpha^l} = \sum_t \delta_j^l[t]\tilde{\alpha}_j^l[t] \tag{27}$$

Note here that

$$\tilde{\alpha}_j^l[t] = \frac{\partial U_j^l[t]}{\partial \alpha^l} \tag{28}$$

Next, we differentiate both sides of Eq. (1) with respect to $\alpha^l$, ignoring the illusory component originated from the surrogate gradient. Thus, we get:

$$\frac{\partial U_j^l[t]}{\partial \alpha^l} = \alpha^l \frac{\partial U_j^l[t-1]}{\partial \alpha^l} + (U_j^l[t-1] - \theta_j^l s_j^l[t-1]) \tag{29}$$

which actually means:

$$\tilde{\alpha}_j^l[t] = \alpha^l \tilde{\alpha}_j^l[t-1] + (U_j^l[t-1] - \theta_j^l s_j^l[t-1]) \tag{30}$$

This reveals that the leakage trace $\tilde{\alpha}_j^l[t]$ also functions as a trace capable of online evaluation forward in the temporal domain. It is initialized to $\tilde{\alpha}_j^l[0] = 0$, charged by the membrane residual of the corresponding neuron after its reset at each time-step, and decays with the leakage factor.

*Short summary for leakage learning:* Eq. (11), (16), (27), (30) constitute the trace-based leakage learning rule. Its detailed steps along with weight and threshold updates will be described soon in the next subsection. Remember that the updated leakage factors need to be truncated back to be within the range [0, 1].

### *F. Unified Trace-based STOP Synergistic Learning*

We now combine the above trace-based learning of synaptic weights, firing thresholds and leakage factors into our unified memory-efficient temporally-forward trace-based synergistic learning framework for deep feedforward SNNs, wherein the spatial neuron errors can be reused for updates of all the three types of parameters. To learn one sample, at each time-step $t$ = 1, 2, ..., $T$ during the forward pass, the membrane potentials $U_j^l[t]$, as well as the weight-, threshold-, and leakage-related traces $\tilde{w}_{ji}^l[t]$, $\tilde{\theta}_j^l[t]$, and $\tilde{\alpha}_j^l[t]$ of all neurons are computed according to Eq. (1), (20), (26), (30), respectively, in a layer-wise fashion from $l$ = 1 to $l$ = $L$. And the instantaneous loss $E[t]$ is calculated via Eq. (5) or (6) thereafter. Then, the gradient of such loss is spatially backpropagated through all layers down to

**Algorithm 1** STOP: temporally-forward trace-based weight-threshold-leakage synergistic learning for deep SNNs

**Input:** $s_i^0[t]$ — sample-encoded input spike trains;
$s_j^d$ — one-hot desired spike data for output neurons.
**Parameter:** $T$ — total number of time-steps;
$L$ — SNN layer depth;
$N^l$ — number of neurons in layer $l$;
$\eta_w$, $\eta_\theta$, $\eta_\alpha$ — learning rates.
**Output:** $w_{ji}^l$ — updated synaptic weights;
$\theta_j^l$— updated firing thresholds;
$\alpha^l$—updated leakage factors.

Initialize all variables $U_j^l[0]$, $\tilde{w}_{ji}^l[0]$, $\tilde{\theta}_j^l[0]$ and $\tilde{\alpha}_j^l[0]$ to 0;
**for** $t$ = 1 to $T$ **do**
  // Spatial forward
  **for** $l$ = 1 to $L$ **do**
    **for** $j$ = 1 to $N^l$ **do**
      Compute $U_j^l[t]$ by Eq. (1);
      Compute $s_j^l[t]$ by Eq. (2);
      **for** $i$ = 1 to $N_{l-1}$ **do**
        Compute $\tilde{w}_{ji}^l[t]$ by Eq. (20);
      **end for**
      Compute $\tilde{\theta}_j^l[t]$ by Eq. (26);
      Compute $\tilde{\alpha}_j^l[t]$ by Eq. (30);
    **end for**
  **end for**
  // Spatial backward
  **for** $l$ = $L$ to 1 **do**
    **for** $j$ = 1 to $N^l$ **do**
      **if** $l$ = $L$ **then**
        Compute $E[t]$ by either Eq. (5) or (6);
        Compute $\delta_j^L[t]$ by Eq. (11);
      **else**
        Compute $\delta_j^l[t]$ by Eq. (16);
      **endif**
      accumulate $\delta_j^l[t]\tilde{w}_{ji}^l[t]$ to $\Delta w_{ji}^l$ in Eq. (17);
      accumulate $\delta_j^l[t]$ ($\tilde{\theta}_j^l[t]$ - 1) to $\Delta\theta_j^l$ in Eq. (21);
      accumulate $\delta_j^l[t]\tilde{\alpha}_j^l[t]$ to $\Delta\alpha_j^l$ in Eq. (27);
    **end for**
  **end for**
**end for**
// Per-layer averaging for leakage factors
**for** $l$ = 1 to $L$ **do**
  $\Delta\alpha^l \leftarrow \sum_j \Delta\alpha_j^l / N^l$;
**end for**
// Parameter updates
**for** $l$ = 1 to $L$ **do**
  **for** $j$ = 1 to $N^l$ **do**
    **for** $i$ = 1 to $N^{l-1}$ **do**
      Update weight: $w_{ji}^l \leftarrow w_{ji}^l - \eta_w \Delta w_{ji}^l$;
    **end for**
    Update threshold: $\theta_j^l \leftarrow \max(\varepsilon, \theta_j^l - \eta_\theta \Delta\theta_j^l)$;
    //$\varepsilon$ is a small positive numeric
  **end for**
  Update leakage: $\alpha^l \leftarrow \max(0, \min(1, \alpha^l - \eta_\alpha \Delta\alpha^l))$;
**end for**

$l$ = 1, to evaluate the neuron errors $\delta_j^l[t]$ by Eq. (11) and (16). Next, for each neuron, the combinations of its error $\delta_j^l[t]$ and the traces $\tilde{w}_{ji}^l[t]$, $\tilde{\theta}_j^l[t]$, $\tilde{\alpha}_j^l[t]$ at time $t$, i.e., $\delta_j^l[t]\tilde{w}_{ji}^l[t]$, $\delta_j^l[t](\tilde{\theta}_j^l[t]$ - 1), $\delta_j^l[t]\,\tilde{\alpha}_j^l[t]$, are accumulated to corresponding parameter change amounts $\Delta w_{ji}^l$ , $\Delta\theta_j^l$, $\Delta\alpha_j^l$, respectively. Finally, when all time-steps of the training sample are over, these change amounts are scaled by their respective learning rates $\eta_w$, $\eta_\theta$, $\eta_\alpha$ before they are applied to adjusting the corresponding parameters. In this learning procedure, we neither resort to temporally backward computing nor necessitate memorizing all neural states across every time-step in the forward pass, which is memory-costly as in the original BPTT and STBP algorithms. A more precise description of the above trace-based synergistic learning details on one training sample is given in Algorithm 1. Note that the update amounts of leakage factors are per-layer averaged, as all neurons in one layer share the same leakage attribute. Moreover, in convolutional layers, the update amounts of firing thresholds of neurons in each channel should further be averaged if the same firing threshold is adopted by all neurons in one channel [48]. Besides, the updated thresholds have to be truncated above 0 and the updated leakage factors need to be truncated within the range [0, 1], as mentioned in previous subsections.

Fig. 3 illustrates the error/gradient propagation flows in the proposed SNN training algorithm with totally $L$ = 4 layers and $T$ = 4 time-steps unrolled along the temporal axis. Note that for clarity, we only exhibit one neuron at each layer. As can be seen, the temporal gradients, i.e., the weight-, threshold- and leakage-related traces, are propagated forward in the temporal domain, while the spatial gradients, i.e., neuron errors, are propagated backward in the spatial domain. Although the two types of gradients must cooperate at each neuron node to accomplish parameter updates via Eq. (17), (21), (27), they are actually flowing independently of each other without fusion, so that temporal traces never carry information of spatial errors during their propagation, and vice versa. In other words, the spatial and temporal gradients are propagated in orthogonal pathways. This is why our algorithm is called *Spatiotemporal Orthogonal Propagation* (*STOP*). Such paradigm not only saves substantial memory resources thanks to the employment of temporally forward traces, but also greatly simplifies computational graphs compared with the STBP algorithm [36], which has to re-merge spatiotemporally backpropagated error gradients at every neuron and every time-step.

## G. Complexity Analysis

TABLE I further compares the memory and computation complexities of the STBP [36] and our STOP algorithm, where STOP-W refers to the non-synergistic version of the STOP rule that only trains weights, while STOP-WTL refers to the fully synergistic STOP rule that trains weights as well as thresholds and leakages. Without loss of generality, we assume a total of $L$ layers in a fully-connected (FC) SNN, $N$ neurons per layer on average and a sample presentation window of $T$ time-steps long. Since the inference process always keeps the same regardless of particular learning rule, we only count in the memory and compute overheads needed for learning. Further, we merely consider multiplications as computation overheads and ignores other simple operations like additions, as the multiplications would consume much more power on edge systems.

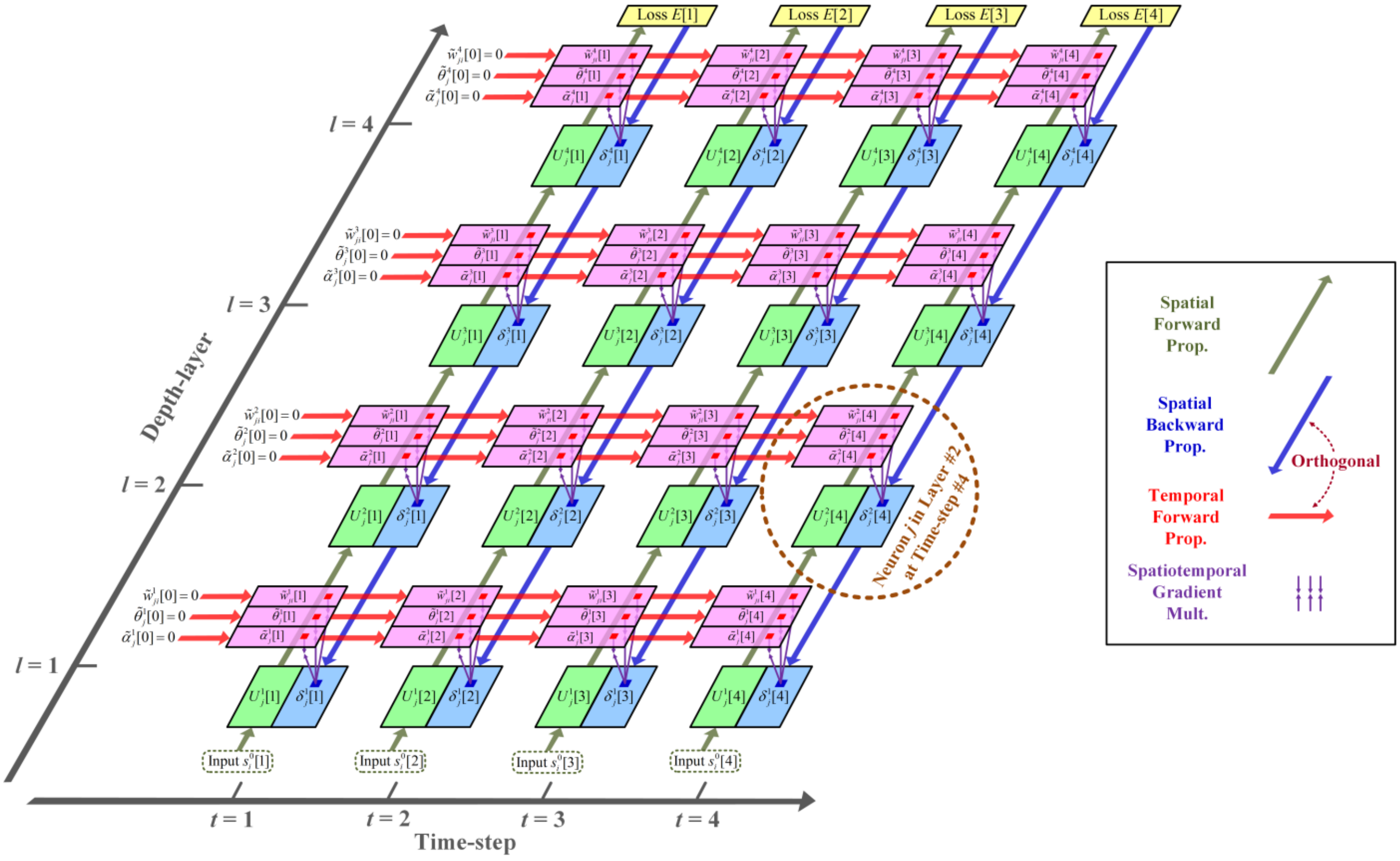

Fig. 3. Orthogonal propagation of spatially backward errors and temporally forward gradients.

*a) STBP:* As STBP requires backpropagating error gradients along both spatial and temporal dimensions, it has to store two neural state variables (i.e., the membrane potential and the fired spike) per neuron per time-step during the forward pass, for later inquiry during the backward pass. Thus, its memory overhead or complexity is $O(2\times N\times L\times T) = O(2NLT)$. Based on the spatiotemporal backpropagation calculation details given in Eq. (18) and (20) in [36], the STBP algorithm calls for $(N+7)$ multiplications to obtain the error of one neuron at a particular time-step from its downstream higher-layer neurons of the same time-step as well as from the neuron itself of the next time-step, and another $N$ multiplications to calculate the updates of its $N$ fan-in synaptic weights at that time-step. Hence, the whole SNN requires $O((N+7+N)\times N\times L\times T)= O((2N+7)NLT)$ multiplications throughout all $T$ time-steps for one training sample.

*b) STOP-W:* Using temporally forward traces, the STOP-W rule has to store the membrane potential, fired spike and the weight-related traces of the neurons only for spatially backward error propagation at current time-step, which are then dropped at the next time-step. As explained above in Section III-C, such weight-related traces are shared by the neurons in one layer, and the number of these traces used by one layer equals to the count of neurons in its preceding layer. Since we have assumed that each layer holds $N$ neurons on average, there is one such trace per neuron. Therefore, the memory complexity for the whole SNN is $O(3\times N\times L) = O(3NL)$. The memory complexity ratio between STBP and STOP-W is $2NLT / 3NL = 2T/3$. In recent researches, $T$ usually falls in the range from 6 to 10 [46], [50], so STOP-W outperforms STBP by at least 4 times in terms of memory reduction. On the other side, for every neuron at each single time-step $t$, STOP-W requires $(N+1)$ multiplications to calculate the spatial neuron errors via Eq. (16), 1 multiplication to update a weight-related trace driven by this neuron's fired spikes and shared by the downstream neurons in the next layer) according to Eq. (20), as well as $N$ multiplications to obtain the update amounts of the neuron's $N$ synaptic weights for time $t$, as indicated by Eq. (17). Therefore, the STOP-W computation complexity for the whole SNN with $NL$ neurons throughout all $T$ times-steps is $O((N+1+1+N)\times N\times L\times T) = O((2N+2)NLT)$, which is approximate to that of STBP, as $(2N+7)NLT) / (2N+2)NLT = (2N+7) / (2N+2) \approx 1$ considering that $N$ is usually at the order of hundreds to thousands. Although it implies that STOP-W and STBP has similar computation complexities, the orthogonal graph of independent spatial and temporal gradient propagations in the STOP-W rule would facilitate customized cost-efficient hardware circuit design.

*c) STOP-WTL:* On top of STOP-W, the fully synergistic learning rule STOP-WTL needs additional memory space for one threshold-related trace and one leakage-related trace per neuron. As a result, its total memory complexity is $O(5\times N\times L) = O(5NL)$, which is still ×1.2 ~ ×2 less than the STBP algorithm that only allows to train synaptic weights, when $T$ varies between 6 and 10 as mentioned above. Likewise, for each neuron, due to the additional 4 multiplications for tracking the threshold- and leakage-related traces in Eq. (26), (30) as well as computing the update amounts of firing thresholds and leakage factors at every time-step $t$ in Eq. (21), (27), the computation complexity slightly rises to $O((N+1+1+N+4)\times N\times L\times T) = O((2N+6)NLT)$. This is almost the same as that of STBP, but our STOP rule can achieve fully synergistic learning towards higher SNN accuracy, and is more cost-effective for edge hardware implementation as analyzed just now.

TABLE I
MEMORY AND COMPUTATION OVERHEADS IN DIFFERENT SNN LEARNING RULES

| Method | Memory complexity | Computation complexity |
|---|---|---|
| STBP [34] | $O(2TLN)$ | $O(TLN(2N+7))$ |
| STOP-W | $O(3LN)$ | $O(TLN(2N+2))$ |
| STOP-WTL | $O(5LN)$ | $O(TLN(2N+6))$ |

Note: $L$ — layer depth, $N$ — average number of neurons per layer, $T$ — number of time-steps for one sample presentation

## IV. EXPERIMENTAL RESULTS

### A. Experimental Setup

In this section, we evaluated the proposed STOP algorithm on both static image datasets (CIFAR-10 [53], CIFAR-100 [53]) and neuromorphic visual datasets (DVS-Gesture [54] and DVS-CIFAR10 [55]) with adequately intermediate-scale deep SNN architectures, including VGG-11 (64C3-P2-128C3-P2-256C3-256C3-P2-512C3-512C3-P2-512C3-512C3-P2-4096-4096-$N_c$, where $N_c$ corresponds to number of object categories in each dataset) and Resnet-18 network structures. As SNNs and our algorithm primarily target on edge intelligence, much larger datasets like ImageNet or very deep networks beyond ResNet-18 were not considered, similar to [36], [39], [43], [44]. During training, we adopted the surrogate gradient $\varphi_{SG}(x) = e^{-|x|}$ for the CIFAR-10 images, while $\varphi_{SG}(x) = 1/(1+\pi^2x^2)$ for others. They have been depicted above in Fig. 2. Throughout our subsequent experiments, the firing thresholds and neural leakages were initialized to 1 and $e^{-1}$ for all neurons whether they participated the synergistic learning or not, and all synaptic weights were randomly initialized following a standard Gaussian distribution (i.e., $\mu = 0$, $\sigma^2 = 1$). More simulation configurations including initial learning rates and weight decay for each case are detailed in the supplementary document.

We employed the direct input coding scheme described in Section II-B to convert pixels of image samples to spike trains. The number of time-steps for one sample presentation were set to 6 for CIFAR-10 and CIFAR-100 datasets. To handle the neuromorphic samples in a unified way with the direct input coding and small number of time-steps, each spike stream in the DVS-CIFAR10 and DVS-Gesture datasets were segmented into multiple slices, each of which contained the same number of successive spike events. Then, the spikes were histogrammed pixel-wisely in each slice. At every time-step $t$, the histogrammed spike counts in the $t$-th slice were fed as a pseudo-frame to the SNN [38], [46], [51]. The number of slices (or equivalently, time-steps) for DVS-Gesture and DVS-CIFAR10 spike streams were set to 20 and 10, respectively.

The learning epochs were set to 200 for each of the datasets. Throughout all subsequent experiments unless stated otherwise, we kept to utilize: 1) the basic stochastic gradient descent (SGD) optimizer [56] with a momentum of 0.9, 2) a cosine annealing scheduler for learning rate [46], and 3) a batch-size of 128. All experiments were conducted on an NVIDIA RTX 3090 GPU.

Furthermore, to conduct ablation study in the effect of our synergistic learning, four variants of the STOP algorithm with different combinations of learnable parameter types were simulated to compare their learning accuracies. These variants included the basic non-synergistic rule STOP-W, the weight-threshold synergistic rule STOP-WT, the weight-leakage synergistic rule STOP-WL, and the fully synergistic rule STOP-WTL. Notably, for convolutional layers, all neurons in each channel share one firing threshold [48], [50]. Therefore, threshold update amounts calculated by the neurons in the same channel should be averaged before being applied to updating their shared threshold in Line 40 of Algorithm 1.

### B. Validation Results

Fig. 4 exhibits the simulation results of our STOP algorithm with different synergistic learning policies. It reports accuracy metrics and learning latencies across various datasets and SNN architectures with both CE and MSE loss functions. Actually, we have the following key observations from two aspects.

*Non-synergistic vs. synergistic learning:* In Fig. 4, visual classification accuracies obtained by synergistic learning rules exceeded the accuracies attained by non-synergistic ones in almost all cases, yet at the cost of increased learning latencies. Particularly, when using the CE loss, the synergistic STOP-WT and STOP-WL achieved similar accuracies, which are higher than the non-synergistic STOP-W but lower than the fully synergistic STOP-WTL. These accuracy discrepancies between non-synergistic and fully-synergistic STOP also appeared when switching to the MSE function, as indicated in Fig. 4. Especially, the accuracy significantly increased by an amount of 1.86% from STOP-W to STOP-WTL on the challenging CIFAR-100 + ResNet-18 SNN benchmark with the MSE loss. These experimental results successfully validated the efficacy of synergistic learning for deep SNNs. However, high performance of synergistic learning comes at a cost of higher learning latencies, as shown in Fig. 4. This is in accordance with the memory complexity analysis given in Table I, as higher memory complexity implies more time consumptions on memory accesses. The fully synergistic learning required several to around 20 more seconds than the non-synergistic baseline to accomplish one training epoch on diffident datasets with SNNs of varying depths. Such elongated learning time would be unacceptable in some strictly real-time or high-speed scenarios. Consequently, a careful tradeoff between accuracy performance and learning latency is necessary when choosing proper synergistic strategies for real applications.

*CE loss vs. MSE loss*: When applying either non- or fully-synergistic learning to the benchmarks, the MSE loss function always behaved marginally poorly in terms of classification accuracy compared with the CE loss, exhibiting a degradation from -0.29% on the simple CIFAR-10 + VGG-11 case to -2.26% on the more complex CIFAR-100 + ResNet-18 task using non-synergistic learning. This gap reduced to -0.3% ~ -1.73% under the fully synergistic learning framework. On the other hand, the difference of learning latencies between using CE and MSE losses with an identical synergistic strategy varied only on a sub-second scale when simulated in software on the GPU processor, as shown in Fig. 4. Nevertheless, when the learning procedures are required to run on customized neuromorphic hardware to enable on-chip self-adaptation, the complicated exponential and division operations required by the SoftMax function as given in Eq. (7) would consume a considerable amount of hardware resources [57]. So, dedicate neuromorphic hardware processors might prefer the simpler MSE loss to the CE loss unless on-chip learning accuracy is extremely pursued.

### C. Work Comparison and Discussion

We have further compared the proposed STOP algorithm with prior state-of-the-art deep convolutional SNN learning methods in terms of object recognition accuracy in Table II. Evidently, on each dataset, when using similar SNN structures and roughly the same count of time-steps, our STOP algorithm, especially the fully synergistic STOP-WTL rule, outperforms most of other spatiotemporal gradient-based training methods. It is noteworthy that the SLTT algorithm in [46] leverages an additional step called weight standardization (i.e., Gaussian normalization to $\mu = 0$ and $\sigma^2 = 1$) to improve accuracy at the cost of higher computational overheads. Although it slightly exceeds our accuracies on the DVS-CIFAR10 dataset with a VGG-11 SNN, it lags behind our metrics on all other datasets (i.e. CIFAR-10, CIFAR-100, DVS-Gesture) using either VGG-11 or Resnet-18 SNN. The prefix *NF-* in their network names just indicates the weight standardization manipulations and does not imply structural modification [58]. Interestingly, it can be found in Table II that on the CIFAR-10 dataset, the VGG-11 SNN obtains accuracies inferior to that of the CIFARNet and Net2 structures holding 2-3 less layers. We deduce this may attributed to the fact that the deeper VGG-11 requires as more as 5 pooling layers, and this leads to a total collapse of spatial information on the low-resolution 32×32 CIFAR-10 images before entering FC layers for classification. In contrast, the CIFARNet and Net2 SNNs have only 2 pooling layers, thus preserving sufficient 8×8 spatial clues and allowing for better classification results in FC layers. However, if more properly deeper structures are employed, such as the Resnet-18 which has a much deeper depth with residual connections, the CIFAR-10 recognition accuracy can be greatly improved, as shown in Table II. On the other side, it is observed in Table II that the works of [59] and [38] using the powerful ResNet structures obtain much lower accuracies than those employing the VGG-11 on the DVS-CIFAR10 set. The reason lies in their adopted training/testing set partition different from that of other works.

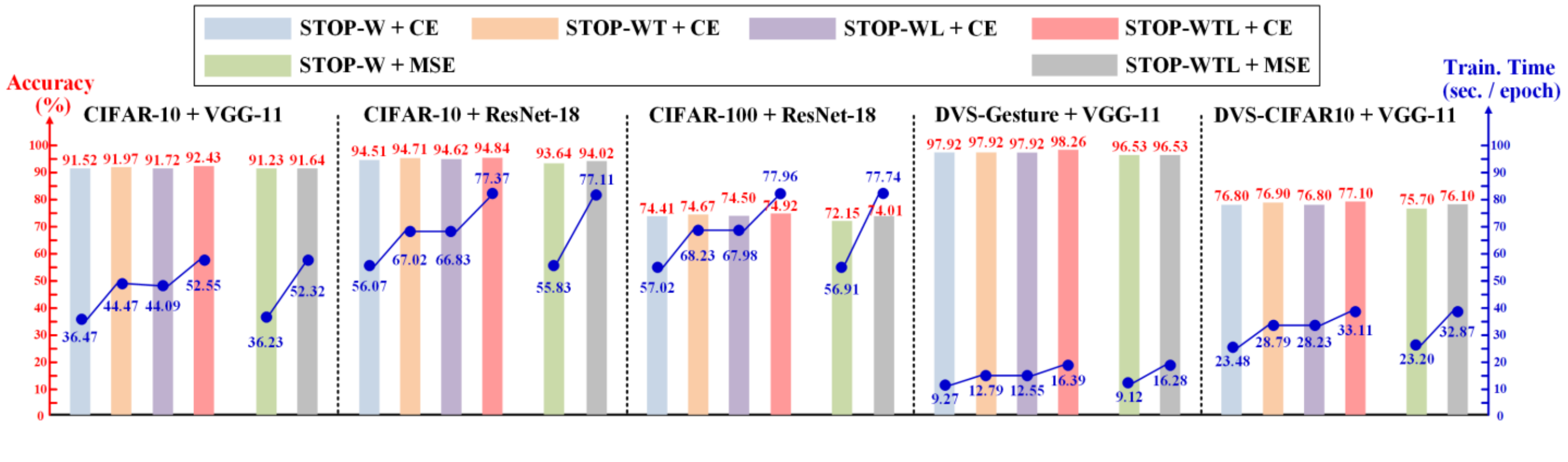

Fig. 4. Learning accuracy and latency over various datasets and SNN architectures with CE and MSE loss functions.

TABLE II

RELATED WORK COMPARISONS IN TERMS OF OBJECT RECOGNITION ACCURACIES.

| Dataset | Method | SNN Network Architecture | Time-steps | Synergistic Learning | Accuracy (%) |
|---|---|---|---|---|---|
| CIFAR-10 | EIHL [59] | VGG-11 | *T* = 6 | No (-W) | 85.75% |
| | **STOP-W (ours)** | **VGG-11** | ***T* = 6** | **No (-W)** | **91.52%** |
| | **STOP-WTL (ours)** | **VGG-11** | ***T* = 6** | **-WTL** | **92.43%** |
| | STBP + NeuNorm [37] | CIFARNet [1] | *T* = 8 | No (-W) | 90.53% |
| | Improved STBP [48] | CIFARNet [1] | *T* = 8 | -WT | 89.40% |
| | **STOP-W (ours)** | CIFARNet [1] | ***T* = 5** | **No (-W)** | **91.75%** |
| | **STOP-WTL (ours)** | CIFARNet [1] | ***T* = 5** | **-WTL** | **92.34%** |
| | ELL [43] | Net2 [2] | *T* = 10 | No (-W) | 89.40% |
| | BELL [43] | | | | 88.01% |
| | FELL [43] | | | | 88.36% |
| | STL-SNN [49] | Net2 [2] | *T* = 8 | -WT | 92.42% |
| | PLIF [51] | Net2 [2] | *T* = 8 | -WL | 93.50% |
| | **STOP-W (ours)** | **Net2** [2] | ***T* = 8** | **No (-W)** | **93.55%** |
| | **STOP-WTL (ours)** | **Net2** [2] | ***T* = 8** | **-WTL** | **93.87%** |
| | STBP + tdBN [38] | ResNet-19 [3] | *T* = 6 | No (-W) | 93.16% |
| | TET [40] | ResNet-19 [3] | *T* = 6 | No (-W) | 94.50% |
| | SLTT [46] | NF-ResNet-18 [4] | *T* = 6 | No (-W) | 94.44% |
| | EIHL [59] | ResNet-18 | *T* = 6 | No (-W) | 90.25% |
| | **STOP-W (ours)** | **ResNet-18** | ***T* = 6** | **No (-W)** | **94.51%** |
| | **STOP-WTL (ours)** | **ResNet-18** | ***T* = 6** | **-WTL** | **94.84%** |
| CIFAR-100 | TET [40] | ResNet-19 [3] | *T* = 6 | No (-W) | 74.72% |
| | SLTT [46] | NF-ResNet-18 [4] | *T* = 6 | No (-W) | 74.38% |
| | EIHL [59] | ResNet-18 | *T* = 6 | No (-W) | 58.63% |
| | STL-SNN [49] | ResNet-18 | *T* = 4 | -WT | 72.87% |
| | **STOP-W (ours)** | **ResNet-18** | ***T* = 6** | **No (-W)** | **74.41%** |
| | **STOP-WTL (ours)** | **ResNet-18** | ***T* = 6** | **-WTL** | **74.92%** |
| DVS-Gesture | STL-SNN [49] | 128C7-P2-128C3-P2-128C3-P2-128C3-P2-128C3-P2 -512-11 | *T* = 20 | -WT | 97.22% |
| | STBP + tdBN [38] | ResNet-17 [5] | *T* = 40 | No (-W) | 96.87% |
| | SLTT [46] | NF-VGG-11 [4] | *T* = 20 | No (-W) | 97.92% |
| | PLIF [51] | VGG-11 | *T* = 20 | -WL | 97.57% |
| | **STOP-W (ours)** | **VGG-11** | ***T* = 20** | **No (-W)** | **97.92%** |
| | **STOP-WTL (ours)** | **VGG-11** | ***T* = 20** | **-WTL** | **98.26%** |
| DVS-CIFAR10 | STBP + NeuNorm [37] | 128C3-128C3-P2-384C3-384C3-P2-1024-512-10 | *T* = 10 | No (-W) | 60.50% |
| | STL-SNN [49] | 128C7-P2-128C3-P2-128C3-P2-128C3-P2-1024-1024-10 | *T* = 20 | -WT | 77.30% |
| | EIHL [59] | ResNet-18 | *T* = 6 | No (-W) | 62.90% |
| | STBP + tdBN [38] | ResNet-19 [3] | *T* = 10 | No (-W) | 67.80% |
| | SLTT [46] | NF-VGG-11 [4] | *T* = 10 | No (-W) | 77.17% |
| | EIHL [59] | VGG-11 | *T* = 6 | No (-W) | 62.45% |
| | PLIF [51] | VGG-11 | *T* = 20 | -WL | 74.80% |
| | TET [40] | VGG-11 | *T* = 10 | No (-W) | 76.30% |
| | **STOP-W (ours)** | **VGG-11** | ***T* = 10** | **No (-W)** | **76.80%** |
| | **STOP-WTL (ours)** | **VGG-11** | ***T* = 10** | **-WTL** | **77.10%** |

[1] CIFARNet: 128C3-256C3-P2-512C3-P2-1024C3-512C3-1024-512-10.
[2] Net2: 256C3-256C3-256C3-P2-256C3-256C3-256C3-P2-2048-100-10
[3] ResNet-19: ResNet-18 extended with one extra FC layer
[4] NF means using weight standardization
[5] ResNet-17: ResNet-18 with the first FC layer removed

Particularly, the works in [38] and [59] have used only 8000 samples for training and left 2000 samples for testing, while other researches have taken 9000 samples for training and left 1000 samples for testing. Consequently, the works [38] and [59] suffer a lower accuracy on the DVS-CIFAR10 due to fewer training samples. Such lack of training samples could not be compensated by the ResNet structures, or possibly even incurs overfitting issues and further degrades the accuracy when using complicated SNNs.

We have argued earlier in Section III-C that temporal flows of illusory spatial gradient components caused by SG functions (as done in the STBP algorithm and its variants) would be ineffective or even detrimental to the accuracy performance. They are removed from our trace-based STOP framework. For a fair validation of such hypothesis without interference from other less relevant elements, we re-implemented the baseline STBP method [36] in PyTorch to evaluate its learning accuracy on the benchmark datasets under totally the same settings as we configured in Section IV-A. Fig. 5 clearly illustrates that our non-synergistic STOP-W rule significantly outperforms the basic STOP in terms of learning accuracy across all benchmarks. In Table II, the ELL [43], FELL [43] and SLTT [46] algorithms are also executed in a temporally-forward manner without propagating the illusory SG gradients during learning, but they do not perform synergistic learning on firing thresholds or leakage parameters to further boost their recognition accuracies. On the other hand, although the works [48], [49], [51] in Table II achieve partially synergistic learning, they do not preclude the illusory components during their temporal backpropagation. Hence, their accuracies are not maximized. In comparison, to the best of our knowledge, our STOP framework is the first one that manages to combine advantages from both temporally-forward learning procedure and synergistic learning paradigm, leading to high accuracies on challenging benchmarks without any aid of complicated optimization techniques. Our STOP framework allows a flexible choice from non- to fully-synergistic learning for deep SNNs, and thus enables tradeoffs among hardware cost and recognition accuracy upon practical demands. These characteristics make our work very attractive for resource-constrained edge intelligent applications.

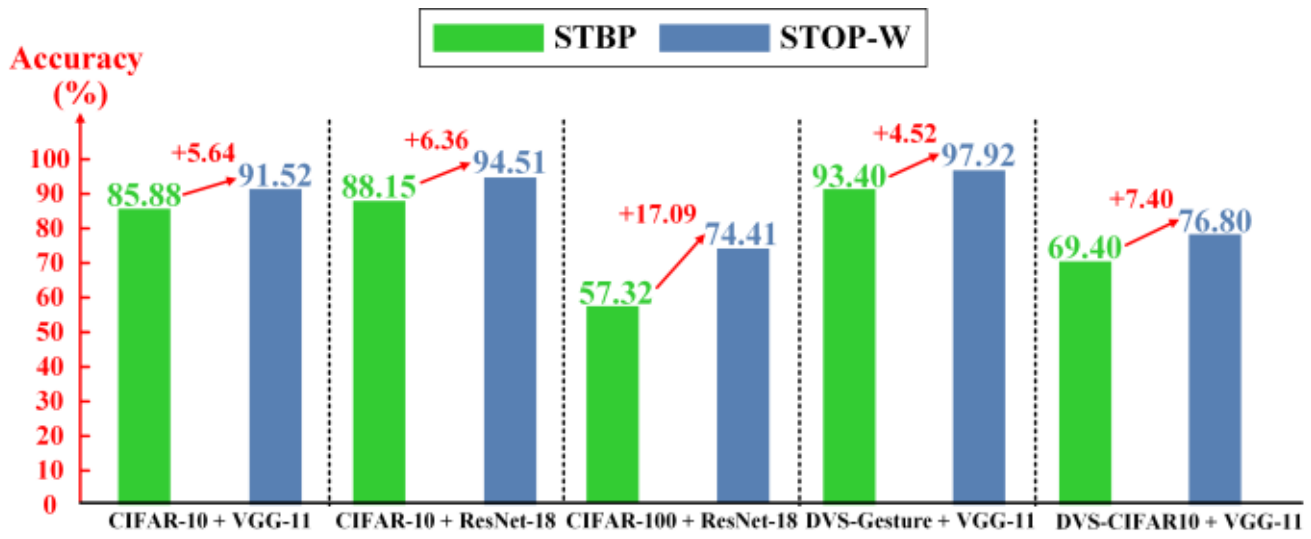

Fig. 5. Deep SNN learning accuracy comparison between the vanilla STBP algorithm [36] and the proposed STOP algorithm.

## V. Conclusion

This work proposes the STOP algorithm for high-accuracy low-memory-complexity training of deep feedforward SNNs. To achieve this goal, our algorithm enables weight-threshold-leakage synergistic learning under a unified temporally-forward trace-based learning framework. The employment of temporal traces eliminates the huge memory requirement for storing neuronal states across all time-steps in the forward pass, while the synergistic learning can automatically find optimal firing thresholds and leakage parameters to improve SNN accuracy. Moreover, in our STOP framework, the spatial gradients (i.e., errors) and the temporal gradients (i.e., traces) propagate orthogonally to and independently of each other, substantially reducing computational complexity. Elaborate experiments and work comparisons demonstrate the efficacy of the proposed algorithm on a variety of datasets and deep convolutional SNNs. These results indicate that our method is quite suitable for self-adaptive neuromorphic intelligence at the edge with limited resources.

**Haoran Gao** (S'21) received the B.S. degree in electronic information engineering from Nanjing University of Information Science and Technology, Nanjing, China, in June 2021. He is currently working toward the Ph.D. degree with the School of Microelectronics and Communication Engineering, Chongqing University, Chongqing, China. His research interests include neuromorphic computing algorithms.

**Xichuan Zhou** (S'06–M'13-SM'21) received the B.S. and Ph.D. degrees from Zhejiang University, Hangzhou, China, in 2005 and 2010, respectively. He was a Visiting Scholar with Arizona State University, Tempe, AZ, USA, in 2009. He was the Vice Dean of the School of Microelectronics and Communication Engineering, Chongqing University, in 2022. He is currently the Director of the Institute of Science on Brain Inspired Intelligence, Chongqing University, Chongqing, China. He made original contributions to intelligent edge computing, significantly contributing to both efficient deep learning methods and engineering applications. He has authored or coauthored more than 60 papers in prestigious international journals and conferences, including IEEE TNNLS, IEEE TCAS-I, IEEE TED, IEEE TBioCAS, IEEE SPL, IEEE GRSL, ICML, AAAI, and CVPR. His monograph of Deep Learning on Edge Computing Devices is one of the earliest books covering multidisciplinary topics from algorithm to hardware design of embedded AI systems. He also held over a dozen patents and has given numerous keynote speeches and invited talks, and chaired several conferences. He was awarded the Outstanding Scientist of Chinese Institute of Electronics in 2021.

**Yingcheng Lin** received his B.S. M.S. and Ph.D. degrees in electrical engineering from Chongqing University, Chongqing, China, in 2006, 2009 and 2014, respectively. Since 2017, he is an associate professor with the School of Microelectronics and Communication Engineering, Chongqing University, China. He is engaged in high-performance embedded signal processing system designs.

**Min Tian** (M'17) received her B.S. degree from Sun-Yat Sen University, Guangzhou, China, in 2014, and the Ph.D. degree from the Institute of Microelectronics, Chinese Academy of Sciences, Beijing, China, in 2019. From Oct. 2020 to Aug. 2023, she was a research assistant professor with the School of Microelectronics and Communication Engineering, Chongqing University, Chongqing, China, where she has been a lecturer from Sept. 2023 to Aug. 2024. She is currently a senior engineer with Chongqing United Microelectronics Center Co. Ltd, Chongqing, China. Her research interests include emerging semiconductor technology, devices and circuits for ubiquitous AIoT applications.

**Liyuan Liu** (M'11) received the B.S. and Ph.D. degrees in Electrical Engineering from the Electronic Engineering Department, Institute of Microelectronics, Tsinghua University, China, in 2005 and 2010, respectively. He is currently a full professor with the Institute of Semiconductors, Chinese Academy of Sciences, Beijing, China. His research interests include high-end AI-enabled vision sensors, processors and algorithms.

**Cong Shi** (S'13-M'17) received his B.S. degree in electronic information science and technology, M.S. degree in microelectronics from Harbin Institute of Technology, Harbin, China, in 2007 and 2009, respectively, and the Ph.D. degree in electrical engineering from Tsinghua University, Beijing, China, in 2014. He was also a joint Ph.D. student at the Institute of Semiconductors, Chinese Academy of Sciences, Beijing, China. From 2015 to 2018, he was a postdoctoral fellow with the Schepens Eye Research Institute, Massachusetts Eye and Ear, Harvard Medical School, Boston, MA. He is currently a research professor with the School of Microelectronics and Communication Engineering, Chongqing University, Chongqing China, where he leads a research group in AI chip designs for smart visual processing and neuromorphic systems

# Supplementary Materials

## S1. Experimental Settings

### *A. Dataset and Network Configurations*

Table S1. Detailed learning configurations used in our experiments.

| Dataset | SNN | Time-steps | Synergistic Strategy | Loss | Initial Learning Rate* | Weight Decay |
|---|---|---|---|---|---|---|
| CIFAR-10 | VGG-11 | $T = 6$ | STOP-W | CE | $\eta_w = 1\times10^{-2}$ | $1\times10^{-5}$ |
| | | | STOP-WT | | $\eta_w = 1\times10^{-2}$<br>$\eta_\theta = 2\times10^{-4}$ | |
| | | | STOP-WL | | $\eta_w = 1\times10^{-2}$<br>$\eta_\alpha = 2\times10^{-4}$ | |
| | | | STOP-WTL | | $\eta_w = 1\times10^{-2}$<br>$\eta_\theta = 1\times10^{-4}$<br>$\eta_\alpha = 1\times10^{-4}$ | |
| | | | STOP-W | MSE | $\eta_w = 1\times10^{-2}$ | |
| | | | STOP-WTL | | $\eta_w = 1\times10^{-2}$<br>$\eta_\theta = 1\times10^{-4}$<br>$\eta_\alpha = 1\times10^{-4}$ | |
| | ResNet-18 | $T = 6$ | STOP-W | CE | $\eta_w = 1\times10^{-1}$ | $3\times10^{-4}$ |
| | | | STOP-WT | | $\eta_w = 1\times10^{-1}$<br>$\eta_\theta = 5\times10^{-4}$ | |
| | | | STOP-WL | | $\eta_w = 1\times10^{-1}$<br>$\eta_\alpha = 3\times10^{-4}$ | |
| | | | STOP-WTL | | $\eta_w = 1\times10^{-1}$<br>$\eta_\theta = 3\times10^{-4}$<br>$\eta_\alpha = 1\times10^{-4}$ | |
| | | | STOP-W | MSE | $\eta_w = 1\times10^{-1}$ | |
| | | | STOP-WTL | | $\eta_w = 1\times10^{-1}$<br>$\eta_\theta = 3\times10^{-4}$<br>$\eta_\alpha = 1\times10^{-4}$ | |
| CIFAR-100 | ResNet-18 | $T = 6$ | STOP-W | CE | $\eta_w = 1\times10^{-1}$ | $5\times10^{-4}$ |
| | | | STOP-WT | | $\eta_w = 1\times10^{-1}$<br>$\eta_\theta = 1\times10^{-3}$ | |
| | | | STOP-WL | | $\eta_w = 1\times10^{-1}$<br>$\eta_\alpha = 1\times10^{-3}$ | |
| | | | STOP-WTL | | $\eta_w = 1\times10^{-1}$<br>$\eta_\theta = 5\times10^{-4}$<br>$\eta_\alpha = 5\times10^{-4}$ | |
| | | | STOP-W | MSE | $\eta_w = 1\times10^{-1}$ | |
| | | | STOP-WTL | | $\eta_w = 1\times10^{-1}$<br>$\eta_\theta = 5\times10^{-4}$<br>$\eta_\alpha = 5\times10^{-4}$ | |
| DVS-Gesture | VGG-11 | $T = 20$ | STOP-W | CE | $\eta_w = 1\times10^{-1}$ | $5\times10^{-4}$ |
| | | | STOP-WT | | $\eta_w = 1\times10^{-1}$<br>$\eta_\theta = 1\times10^{-3}$ | |
| | | | STOP-WL | | $\eta_w = 1\times10^{-1}$<br>$\eta_\alpha = 1\times10^{-3}$ | |
| | | | STOP-WTL | | $\eta_w = 1\times10^{-1}$<br>$\eta_\theta = 4\times10^{-4}$<br>$\eta_\alpha = 2\times10^{-4}$ | |
| | | | STOP-W | MSE | $\eta_w = 1\times10^{-1}$ | |
| | | | STOP-WTL | | $\eta_w = 1\times10^{-1}$ | |
| | | | | | $\eta_\theta = 4\times10^{-4}$<br>$\eta_\alpha = 2\times10^{-4}$ | |
| DVS-CIFAR10 | VGG-11 | $T = 10$ | STOP-W | CE | $\eta_w = 5\times10^{-2}$ | $5\times10^{-4}$ |
| | | | STOP-WT | | $\eta_w = 5\times10^{-2}$<br>$\eta_\theta = 4\times10^{-4}$ | |
| | | | STOP-WL | | $\eta_w = 5\times10^{-2}$<br>$\eta_\alpha = 2\times10^{-4}$ | |
| | | | STOP-WTL | | $\eta_w = 5\times10^{-2}$<br>$\eta_\theta = 3\times10^{-4}$<br>$\eta_\alpha = 1\times10^{-4}$ | |
| | | | STOP-W | MSE | $\eta_w = 5\times10^{-2}$ | |
| | | | STOP-WTL | | $\eta_w = 5\times10^{-2}$<br>$\eta_\theta = 3\times10^{-4}$<br>$\eta_\alpha = 1\times10^{-4}$ | |

* Initial learning rate for cosine annealing.